\pdfoutput=1
\documentclass[runningheads]{llncs}
\usepackage{graphicx}
\usepackage{booktabs}
\usepackage{color}
\usepackage{marvosym}
\usepackage{multirow,multicol}
\usepackage{amssymb}
\usepackage{amsmath}
\begin{document}
\title{Large language models improve Alzheimer's disease diagnosis using multi-modality data}
\author{ Yingjie Feng\inst{1} \and
 Jun Wang\inst{1}\ \and
 Xianfeng Gu \inst{2}\and
 Xiaoyin Xu \inst{3} \and
 Min Zhang\inst{4} \textsuperscript{(\Letter)} 
 }

 \authorrunning{Y. Feng et al.}
 \institute{School of Software Technology, Zhejiang University, Hangzhou, China \and
 Department of Computer Science, Stony Brook University, Stony Brook, NY, USA\and
 Department of Radiology, Brigham and Women's Hospital, Harvard Medical School, Boston, MA, USA \and
 College of Computer Science and Technology, Zhejiang University, Hangzhou, China \\ \email{min\_zhang@zju.edu.cn}
 }

\maketitle              
\begin{abstract}
In diagnosing challenging conditions such as Alzheimer's disease (AD), imaging is an important reference. Non-imaging patient data such as patient information, genetic data, medication information, cognitive and memory tests also play a very important role in diagnosis. effect. However, limited by the ability of artificial intelligence models to mine such information, most of the existing models only use multi-modal image data, and cannot make full use of non-image data. We use a currently very popular pre-trained large language model (LLM) to enhance the model's ability to utilize non-image data, and achieved SOTA results on the ADNI dataset.

\end{abstract}

\section{Introduction}
Alzheimer's disease (AD) is a progressive neurodegenerative disorder characterized by the permanent deterioration of memory, language, and cognitive functions in affected individuals~\cite{de2016cellular}. The disease progresses slowly and consists of several stages, eventually leading to significant dementia in about 60-80\% of patients~\cite{gaugler20222022}. There is currently no cure for AD, but early diagnosis of its early stages, namely mild cognitive impairment (MCI), is critical to slowing disease progression and improving patients' quality of life~\cite{ning2021relation}. Existing machine learning methods~\cite{song2022multi,pei2022multi} have shown great success in AD classification problems. However, the existing technology still has many limitations, which are mainly reflected in the multi-modal data integration~\cite{hao2020multi}. First, the data of AD patients usually include various examination data and multimodal image data of brain scans. Previous work such as~\cite{song2021graph} has shown that the combination of image data and other related data is beneficial to the improvement of model performance, but how to efficiently combine statistical non-imaging data and medical image data is still a hot research issue. Second, although non-image data has been used in many papers~\cite{cobbinah2022reducing,song2021graph,pei2022multi}, due to the limitation of the methods,  information provided by non-image data for classification is limited. In many ablation experiments, the presence or absence of non-image data has little impact on classification accuracy. This shows that the existing models have limited understanding of this type of text and tabular data. Traditional feature engineering methods also have limited capabilities in this field. Using a better language model to improve the understanding of the machine will be a possible solution.

Large language models (LLMs), based on the transformer architecture~\cite{vaswani2017attention}, have revolutionized natural language processing, showing remarkable performance in generating and interpreting sequences across various domains, such as natural language, computer code, and protein sequences. The scale of the model, including model size, dataset size, and training computation, has been shown to be crucial for robustness in inferences from large neural models~\cite{liu2023pre}. LLMs also have the potential to make useful inferences for a broad range of specialized tasks without dedicated fine-tuning, including assisting with medical problem solving~\cite{brown2020language}. The recently released GPT-4 model has significantly larger model parameters and training data than GPT-3.5, which is the model behind ChatGPT~\cite{openai2023gpt}. The use of LLMs in medicine has a long-standing research program, with various representations and reasoning methods explored over the decades~\cite{liu2023pre}. 
In the medical domain, LLMs have demonstrated their potential as valuable tools for providing medical knowledge and advice~\cite{nori2023capabilities}. For instance, a large dialog-based LLM such as ChatGPT has demonstrated remarkable results in a critical evaluation of its medical knowledge~\cite{wang2023chatcad}. ChatGPT has successfully passed part of the US medical licensing exams~\cite{singhal2022large}, showcasing its potential to augment medical professionals in delivering care. 

\begin{figure*}[h]
\includegraphics[width=\textwidth]{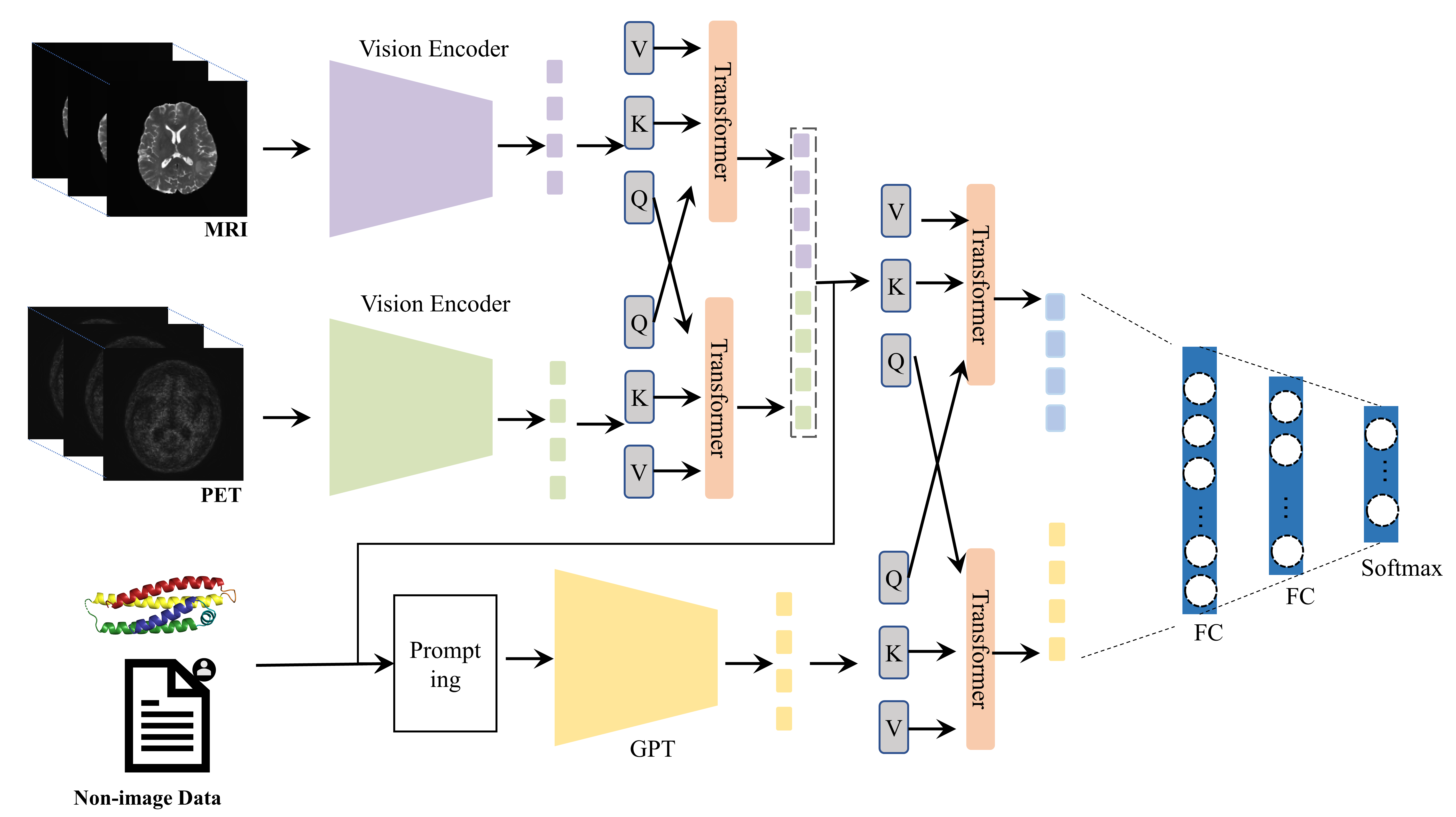}
\caption{Architecture of our network} 
\label{fig1}
\end{figure*}

Based on the above work and ideas, we propose a new model as shown in Fig\ref{fig1}. Building on the general artificial intelligence and multi-modal feature space capabilities brought by LLM, we apply the concept of cross-attention to fuse and align data from different domains, achieving better application of various modal data.
Our main contributions are: 1) applying an LLM to non-image data for knowledge embedding and multimodal alignment, and 2) proving the effectiveness of our method on the ADNI dataset and reaching SOTA level.

\section{Methods}
\subsection{Embedding of Non-image Data}
\begin{figure*}[h]
\includegraphics[width=\textwidth]{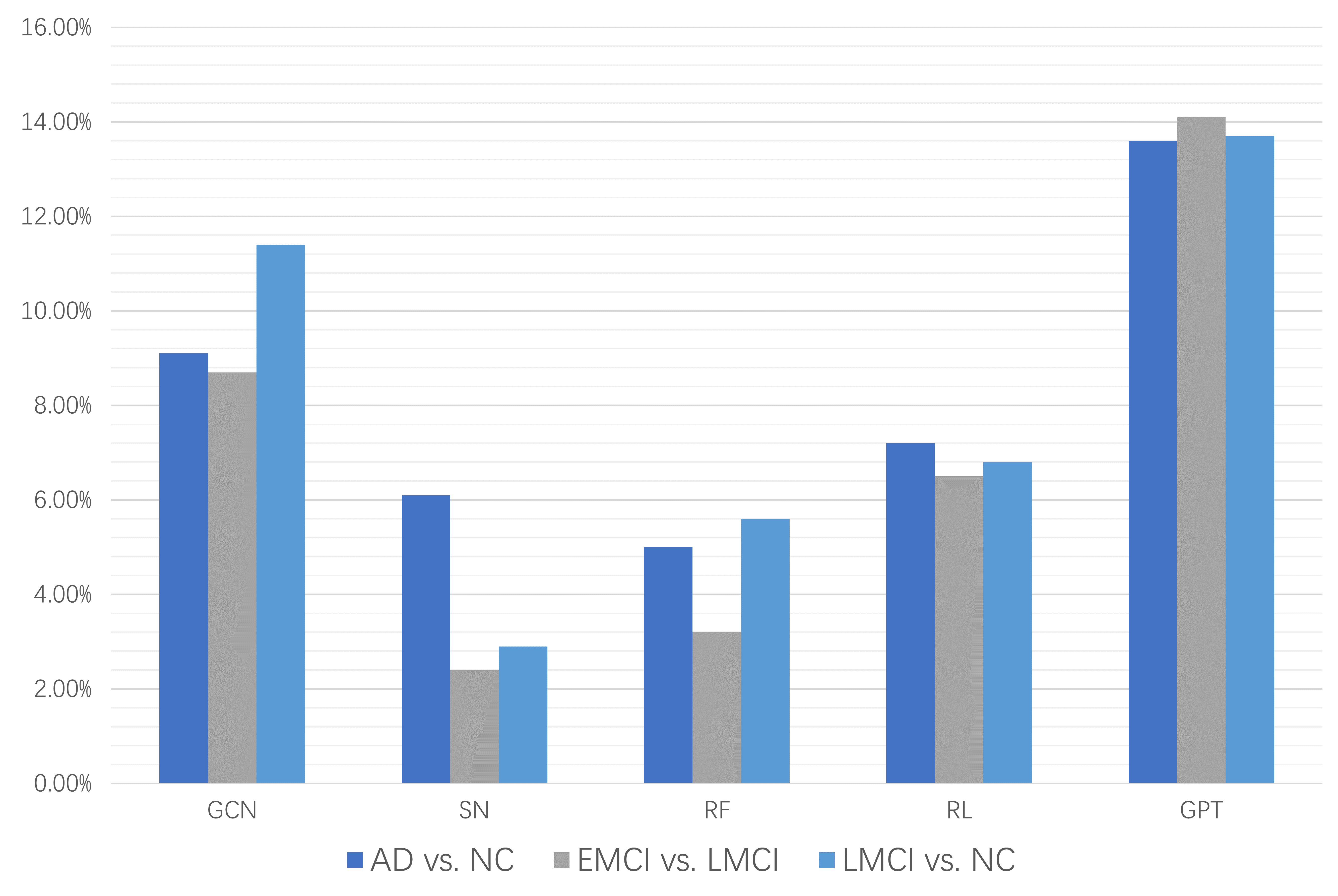}
\caption{Using different non-image data processing methods to improve the performance of the model on three common binary classification tasks of ADNI.} 
\label{fig2}
\end{figure*}

The ADNI data set contains various data types related to patients, such as clinical data, which includes demographic information of subjects (such as gender, age, education level, family medical history, etc.), neuropsychological data, which includes Mini-Mental State Examination Psychological test results and Alzheimer's Disease Assessment Scale-Cognitive Subscale (ADAS-Cog), imaging data, cerebrospinal fluid biomarker data, and genomics data. Most artificial intelligence models based on computer vision will choose to include some of the information to improve the performance of the model. Commonly used ones include genomic information APOE, patient age, cognitive test results MMSE, and cerebrospinal fluid marker A$\beta$.
In order to allow the model to be used effectively, non-image data is often added to the model using various embedding methods. Commonly used embedding methods include: simple normalization (SN), random forest (RF), graph neural network methods (GCN), and representation learning (RL) etc. In our experiments, we compared the improvement of classification performance by these common methods, as shown in Fig.~\ref{fig2}. 
These embedding methods have a certain effect on improving the model, but operations such as data normalization, data processing, and feature selection need to be performed in advance based on pathology and doctor expertise, so that the model can recognize and use various types of information. Therefore, the performance improvement brought by such non-image data to the model is more based on the understanding of the disease by the model designer and doctors rather than the ability of the model. The pre-trained LLM has shown the ability to approach artificial general intelligence (AGI) in many fields, and has many applications in the medical field. GPT-4 can even pass the medical license exam. Therefore, we hope to use GPT to process non-image data, so as to make better use of this information to achieve better classification and diagnosis results.
After showing some examples to GPT, we group various non-image information to input to GPT, and GPT generates the feature tokens of these non-image data to participate in subsequent processing. We compared feature tokens of different dimensions, and we found that the performance of 64 feature values output by GPT is the best.
GPT-4 has added image information in the pre-training, so it has a strong multi-modal fusion ability. But at present, the API of GPT-4 with image input cannot be used normally. So we use image features instead of the image itself to participate in the fusion process with non-image data.

\subsection{Modality Alignment}
In terms of multimodal fusion, we use the cross-attention to concatenation method to fuse PET and MRI images. This method is proposed based on multihead-self-attention and cross-attention~\cite{xu2022multimodal}.
In the architecture of vanilla transformer, the central component is the self-attention (SA) operation, also known as "Scaled Dot-Product Attention"~\cite{vaswani2017attention}. The input sequence $X = [x_1, x_2, · · · ] \in \mathbb{R}^{N×d} $ undergoes optional positional encoding through point-wise summation $Z \gets X \oplus PositionEmbedding$ or concatenation $Z \gets concat(X, P ositionEmbedding)$. After preprocessing, the $Z$ embedding is projected onto three matrices, $W^Q \in \mathbb{R}^{d \times d_q}, W^K \in \mathbb{R}^{d \times d_k},$ and $W^V \in \mathbb{R}^{d \times d_v}$, where $d_q = d_k$, generating the Q (Query), K (Key), and V (Value) embeddings as $$ Q = ZW^Q, K = ZW^K,$$ and $$V = ZW^V.$$ The output of self-attention is defined as $ Z = SA(Q, K, V) = Softmax \left( \frac{QK ^{\top} }{\sqrt{d_q}}\right) V$. Through self-attention, each input element can attend to all the others, encoding the input as a fully-connected graph. Consequently, the encoder of the vanilla transformer can be interpreted as a fully-connected Graph Neural Network (GNN) encoder, and the transformer family possesses the non-local ability of global perception, similar to the non-local network~\cite{xu2022multimodal}.
Multi-Head Self-Attention (MHSA) is a technique in which multiple self-attention sub-layers can be stacked in parallel. Their concatenated outputs are then fused by a projection matrix W, forming a structure known as MHSA. MHSA can be represented as $$Z = MHSA(Q, K, V) = concat(Z_1, \dots, Z_H),$$ where each head $Z_h = SA(Q_h, K_h, V_h) $ with $h \in [1, H] $, and $W$ is a linear projection matrix. The concept behind MHSA is similar to ensemble learning. By using MHSA, the model can attend to information from multiple representation sub-spaces simultaneously, improving its ability to extract relevant information from the input sequence.
The two streams of cross-attention can be concatenated and processed by another transformer $Tf(concat(a,b))$ to model the global context. This approach, known as hierarchically cross-modal interaction, has been widely studied and is used to alleviate the limitation of cross-attention. By concatenating the cross-attention streams, the model can better capture the relationships between the input sequences and their corresponding modalities, resulting in improved performance.
\begin{eqnarray}    
       Z_{(A)}\gets MHSA(Q_B,K_A,V_A), \nonumber \\  
  Z_{(B)}\gets MHSA(Q_A,K_B,V_B), \nonumber\\ 
  Z\gets Tf(C(Z_{(A)},Z_{(B)})). \nonumber 
  \end{eqnarray} 
In the fusion of MRI and PET image modalities, as well as the fusion of image and non-image features, we have used the above-mentioned cross-attention to concatenation method.
In order to allow LLM to refer to image features when performing feature extraction, we input the image features after cross-attention into LLM. Through such a connection form, we have carried out the modal cross operation in the stage of multiple modal fusion, so as to achieve a better fusion effect. Ablation experiments show that this multi-level fusion brings about 3\% performance improvement.

\subsection{Model}
In order to adapt to different characteristics of two images, we use two independent vision encoders for image feature extraction. To avoid overfitting, we choose a network model with a smaller depth. After comparing different models (as shown in Table \ref{tab1}), we finally chose to use ConvNeXt as the vision encoder. Compared with the traditional CNN network, ConvNeXt has absorbed more advantages of the Transformer structure, so it is more suitable for the transformer structure in our framework. Compared with ViT, ConvNeXt retains more convolution operations, making training more effective.

For the classification of the final features, we used a multi-layer perceptron (MLP) structure constructed with three fully-connected layers. In the softmax layer of the final layer, we dynamically adjust the shape of output vector  based on the task, thereby achieving the ability to handle multiple tasks such as binary, ternary, and quaternary classifications on the same model.

\section{Experiments and Results}
\subsubsection{Data Description} We evaluate our Multi-Modal Semi-supervised Evidential Adversarial Network using the Alzheimer’s Disease Neuroimaging Initiative (ADNI) dataset2. ADNI is a multi-center dataset composed of multi-modal data including imaging and multiple phenotype data. 
The dataset contains four categories: early mild cognitive impairment (EMCI), late mild cognitive impairment (LMCI), normal control (NC), and AD. 


\subsubsection{Evaluation}We design a three part evaluation scheme. First, we follow the majority of techniques convention for binary classification comparing AD vs NC, AD vs EMCI,  LMCI vs NC and EMCI vs LMCI. Second, we extended the classification problem to two multi-class setting including the thress classes AD vs NC vs MCI and four classes AD vs NC vs EMCI vs LMCI. We consider this setting, as one of the major challenges in AD diagnosis is to fully automate the task without pre-selecting classes.
For a fair comparison in performance, we ran all techniques under same conditions. Quality check is performed following standard convention in the medical domain: accuracy (ACC), specificity (SPE), and sensitivity (SEN). At the same time we calculated ROC-AUC on three binary classification tasks.

\subsubsection{Results}
Table \ref{tab1} shows the comparison between our model and the current state of art model, as well as the performance of using different vision encoders. Our model has reached the SOTA level on multiple tasks. The comparison with the original ConvNeXt can demonstrate the use of our proposed non-image data and the effectiveness of the multi-modal fusion method. For the performance improvement obtained by using other non-image data embedding methods, we show the results in Fig.~\ref{fig2}.

Table \ref{tab2} shows the performance of our model on a multi-classification task, indicating that our constructed method is well-suited for 2-4 different categories of classification tasks, thereby enabling the model to better meet the requirements of clinical applications. Our model surpasses or is comparable to SOTA in binary, ternary, and quaternary tasks, proving its performance and adaptability.

Table \ref{tab3} discusses the impact of using different large models and prompt methods on performance. In future work, we will use customized LLMs to adapt to the multi-modal medical image space and the knowledge space of the medical field, and continue to improve model performance through prompt engineering.

\begin{table*}[!h]
\centering
\caption{Performance of different models on three typical binary classification problems on ADNI.}\label{tab1}
\begin{tabular}{lc|cc|cc|cc}
\toprule[1pt]
 \multirow{2}{*}{Method}& \multirow{2}{*}{Backbone} &  \multicolumn{2}{|c|}{AD vs. NC}& \multicolumn{2}{|c|}{EMCI vs. LMCI} & \multicolumn{2}{|c}{LMCI vs. NC} \\
  & & ACC & AUC  & ACC & AUC & ACC & AUC \\
\midrule[2pt]
Baseline~\cite{huang2017densely} & DenseNet& 80.53 & 78.26  & 74.10 & 73.49  & 72.05 & 70.34   \\
SOTA~\cite{pei2022multi}&PKG-Net & 94.30 & 93.75  & 92.92 & {\bfseries 93.14}  & 92.05 & 90.25  \\
SRL~\cite{ning2021relation} & -& {\bfseries96.95} & 94.33 & 84.55& 84.03 & 82.64 & 82.03\\
 \hline
ResNet+GPT~\cite{he2016deep} & ResNet50 & 87.50 & 88.12 & 85.82 & 84.20 & 81.53 & 77.60\\
EfficientNet+GPT~\cite{tan2021efficientnetv2} & EffNetV2-M & 92.52 & 90.11 & 88.50 & 84.75 & 90.34 & 87.68\\
ViT+GPT~\cite{dosovitskiy2020image} & ViT-B/32 & 94.71 & 89.83 & 89.47 & 90.35 & 92.50 & 90.83 \\
\hline
ConvNext~\cite{liu2022convnet} & ConvNeXt-S & 83.59 & 85.70 & 81.45 & 84.63 & 81.50 & 84.12   \\
proposed  & ConvNeXt-S & 96.36 & {\bfseries 97.09} & {\bfseries94.71} & 93.06 & {\bfseries95.28} & {\bfseries 92.87 }  \\
\bottomrule[1pt]
\end{tabular}
\end{table*}

\begin{table*}[!h]
\centering
\caption{The performance of models using multi-class output to directly obtain diagnostic results.}\label{tab2}
\begin{tabular}{l|ccc|ccc}
\toprule[1pt]
 \multirow{2}{*}{Method} &  \multicolumn{3}{c}{AD vs. MCI vs. NC}& \multicolumn{3}{c}{AD vs. EMCI vs. LMCI vs. NC}  \\
  & ACC & SPE & SEN & ACC & SPE & SEN  \\
\midrule[2pt]
Baseline~\cite{huang2017densely} & 61.12 & 60.85 & 59.34 & 58.37 & 54.10  & 51.08 \\
 \hline
U-Net~\cite{fan2021unet} & 87.65 & - & - & 86.47 & - & - \\
slice attention module~\cite{huo2022multistage} & 78.90 & 73.33 & 91.10 & 87.50 & 95.60 & 63.33  \\
Hypergraph Diffusion~\cite{aviles2022multi} & 83.75 & 80.64 & 83.07 & 86.47 &78.52 & 82.16  \\ 
Ours+GPT~\cite{he2016deep} & {\bfseries89.05} & {\bfseries87.33} & 89.29 & {\bfseries87.63} & 85.79 & {\bfseries87.25} \\
\bottomrule[1pt]
\end{tabular}
\end{table*}

\begin{table*}[!h]
\centering
\caption{Error rate on different tasks with different prompt.}\label{tab3}
\begin{tabular}{l|lll|lll|}
\toprule[1pt]
 \multirow{2}{*}{Tasks}   & GPT-4 & GPT-4 & GPT-4 & GPT-3.5 & GPT-3.5 & GPT-3.5  \\
    & (5 shot) & (1 shot) &  (0 shot)  & (5 shot) & (1 shot) & (0 shot) \\
\midrule[2pt]
AD vs NC & 3.64 & 5.19 & 11.26 & 9.05 & 13.18 & 15.72     \\
AD vs MCI vs NC & 10.95 & - & - & - & - & -     \\
AD vs EMCI vs LMCI vs  NC & 12.37 & - & - & - & - & -     \\

\hline
\bottomrule[1pt]
\end{tabular}
\end{table*}

\section{Conclusions}
We have developed a new method to embed non-image information using a large language model and successfully achieved SOTA performance on the ADNI-2 dataset with this method. We have designed a new multiple-time multimodal fusion method and used Experiments demonstrate the effectiveness of modality fusion. We demonstrate the ability of large language models such as GPT to improve diagnostic performance.

%
%
%
%
\bibliographystyle{splncs04.bst}
\bibliography{paper.bib}

\end{document}